\documentclass[lettersize,journal]{IEEEtran}
\usepackage{amsmath,amsfonts}
\usepackage{algorithm}
\usepackage{array}

\usepackage{tikz,xcolor}
\usepackage[implicit=false]{hyperref}
\hypersetup{hidelinks,
	colorlinks=true,
	allcolors=black,
	pdfstartview=Fit,
	breaklinks=true}
\definecolor{lime}{HTML}{A6CE39}
\DeclareRobustCommand{\orcidicon}{
\begin{tikzpicture}
\draw[lime, fill=lime] (0,0)
circle[radius=0.13]
node[white]{{\fontfamily{qag}\selectfont \tiny \.{I}D}};
\end{tikzpicture}
\hspace{-2mm}
}
\foreach \x in {A, ..., Z}{%
\expandafter\xdef\csname orcid\x\endcsname{\noexpand\href{https://orcid.org/\csname orcidauthor\x\endcsname}{\noexpand\orcidicon}}
} 
\usepackage{textcomp}
\usepackage{stfloats}
\usepackage{url}
\usepackage{verbatim}
\usepackage{graphicx}
\usepackage{cite}
\usepackage{xcolor}
\usepackage{tabularx}
\usepackage{supertabular}
\usepackage{ctable}
\usepackage{multirow}
\usepackage{subcaption}
\usepackage{combelow}
\usepackage{fixltx2e}
\usepackage[normalem]{ulem}
\usepackage{xcolor, soul}

\sethlcolor{yellow}
\makeatletter

\newcommand{\Rmnum}[1]{\expandafter\@slowromancap\romannumeral #1@}
\makeatother

\usepackage{hyperref}
\usepackage{algpseudocode}

\begin{document}

\title{Deep Predictive Coding with Bi-directional Propagation for Classification and Reconstruction}

\author{Senhui Qiu\hspace{-2.0mm}\orcidA{}\hspace{-1mm},
Saugat Bhattacharyya\hspace{-2.0mm}\orcidB{}\hspace{-1mm},~\IEEEmembership{Member,~IEEE,} Damien Coyle\hspace{-2.0mm}\orcidC{}\hspace{-1mm},~\IEEEmembership{Senior Member,~IEEE,} and Shirin Dora \hspace{-3.5mm}\orcidD{}\hspace{-1mm}
        % <-this % stops a space
% Manuscript received April 19, 2023; revised May 16, 2023.
\thanks{ This work was supported by the Northern Ireland High Performance Computing (NI-HPC) facility funded by the UK Engineering and Physical Sciences Research Council (EPSRC), Grant No. EP/T022175. Damien Coyle is supported by a UKRI Turing AI Fellowship 2021-2025 funded by the EPSRC under Grant number EP/V025724/1. Senhui Qiu is supported by an Ulster University Vice Chancellor’s Research Scholarship.}

\thanks{Senhui Qiu, Saugat Bhattacharyya and Damien Coyle are
with the Intelligent Systems Research Centre, School of Computing, Engineering and Intelligent Systems, Ulster University, BT48 7JL Londonderry, UK (e-mail: \href{mailto:Qiu-S2@ulster.ac.uk}{Qiu-S2@ulster.ac.uk}; \href{mailto:s.bhattacharyya@ulster.ac.uk}{s.bhattacharyya@ulster.ac.uk}; \href{mailto:dh.coyle@ulster.ac.uk}{dh.coyle@ulster.ac.uk}).}

\thanks{Damien Coyle is also with the Bath Institute for the Augmented Human,  University of Bath, BA2 7AY, Bath, UK. }

\thanks{Shirin Dora is with the Department of Computer Science, Loughborough University, LE11 3TU Loughborough, UK (e-mail: \href{mailto:s.dora@lboro.ac.uk}{s.dora@lboro.ac.uk})}
}

%\markboth{Journal of \LaTeX\ Class Files,~Vol.~14, No.~8, January~2023}%
%{Shell \MakeLowercase{\textit{et al.}}: A Sample Article Using IEEEtran.cls for %IEEE Journals}


\maketitle

\begin{abstract}
This paper presents a new learning algorithm, termed Deep Bi-directional Predictive Coding (DBPC) that allows developing networks to simultaneously perform classification and reconstruction tasks using the same weights. Predictive Coding (PC) has emerged as a prominent theory underlying information processing in the brain. The general concept for learning in PC is that each layer learns to predict the activities of neurons in the previous layer which enables local computation of error and in-parallel learning across layers. In this paper, we extend existing PC approaches by developing a network which supports both feedforward and feedback propagation of information. Each layer in the networks trained using DBPC learn to predict the activities of neurons in the previous and next layer which allows the network to simultaneously perform classification and reconstruction tasks using feedforward and feedback propagation, respectively. DBPC also relies on locally available information for learning, thus enabling in-parallel learning across all layers in the network. The proposed approach has been developed for training both, fully connected networks and convolutional neural networks. The performance of DBPC has been evaluated on both, classification and reconstruction tasks using the MNIST and FashionMNIST datasets. The classification and the reconstruction performance of networks trained using DBPC is similar to other approaches used for comparison but DBPC uses a significantly smaller network. Further, the significant benefit of DBPC is its ability to achieve this performance using locally available information and in-parallel learning mechanisms which results in an efficient training protocol. This results clearly indicate that DBPC is a much more efficient approach for developing networks that can simultaneously perform both classification and reconstruction.

\end{abstract}


\section{Introduction}
\IEEEPARstart{D}{eep} neural networks (DNN) such as AlexNet \cite{NIPS2012_c399862d}, GoogLeNet \cite{DBLP:journals/corr/SzegedyLJSRAEVR14}, VGG \cite{simonyan2014very}, and ResNet \cite{he2016deep}, have performed well on computer vision tasks. These performance benchmarks have been achieved using deeper and wider networks with a large number of parameters which also lead to high computational requirements \cite{lachaux2003simple, lim2021time}. Widespread use of edge devices (like mobile phones and drones) has created a necessity for the development of computationally efficient techniques \cite{murshed2021machine, saranirad2021dob} as limited computing available on these devices impedes the deployment of computationally intensive DNNs. Further, most existing DNNs are trained using error-backpropagation (EBP) \cite{rumelhart1986learning, werbos1982applications} which relies on sequential layer-wise transmission of information from the last to first layer in the network during training. This is termed as the weight transport problem \cite{GROSSBERG198723} and severely affects the efficiency of hardware realizations of EBP \cite{crafton2019local}.

Different from EBP, most forms of plasticity observed in the brain rely on locally available information  on a synapse which circumvents the weight transport problem. Local learning techniques also create opportunities for parallelizing learning across deep networks with many layers \cite{millidge2022predictive, whittington2019theories}. This has motivated researchers to utilize biological phenomena for developing alternative learning techniques. \emph{Predictive coding} (PC) \cite{rao1999predictive} has been proposed as a theoretical model of information processing in the brain. PC utilizes locally available information for learning \cite{millidge2022predictive, whittington2019theories, song2020can} which enables parallelization parameter updates across all layers in the network \cite{millidge2022predictive}. 

% sequential backpropagation of error perform sequential backward updates and non-local computation, unlike those that work in the brain. Therefore, these models impose high resource requirements when realized in hardware and are difficult to parallelize at scale compared with the low-energy-consuming human brain \cite{millidge2022predictive}\cite{whittington2019theories}. This has motivated researchers to develop new and efficient learning algorithms inspired by neuroscience. 

% Currently, EBP is generally considered to lack biologically plausible and unrealistic to be implemented in the brain\cite{millidge2022predictive}\cite{whittington2019theories}. This means weights in EBP cannot update based on locally available information. However, learning in the brain (BL) can utilize locally available information to update weights, which is an alternative learning method that can run in parallel and potentially reduce energy consumption \cite{song2020can}. Recently, some researchers have pay attention back to neuroscience-inspired learning. For instance, a neuroscience-inspired learning algorithm with extremely promising properties is \emph{predictive coding} (PC) \cite{rao1999predictive}, which can utilize locally available information to infer representations and update weights. 

The seminal work of Rao and Ballard \cite{rao1999predictive} developed a neural network based implementation of PC that reproduced various phenomena observed in the visual cortex of the brain. The underlying principle of PC is to build generative models by estimating representations that are capable of reconstructing a given input. Each layer in the network generates predictions about representations associated with the previous layers. PC utilizes the gradient of errors in these predictions to update both representations associated with a given layer and the weights in the network. Both representations and weights are updated in parallel across all layers of the network.

It has also been shown that the representations inferred using PC are also suitable for classification \cite{dora2021deep, sun2020predictive}. This has led to the development of PC based approaches that involve training a single DNN to perform both discriminative tasks like classification and generative tasks like reconstructing an input \cite{sun2020predictive}. Such techniques are particularly beneficial for edge devices as a single network could perform multiple tasks simultaneously. However, most existing algorithms involving PC have utilized locally available information to update the weights for either image classification \cite{millidge2020relaxing, song2018fast} or reconstruction \cite{ororbia2022convolutional} but not both at the same time.

In this paper, we develop a new method called Deep Bi-directional Predictive Coding (DBPC) which can be used to build networks that can simultaneously perform classification and reconstruct a given input. The networks trained using DBPC are referred to as Deep Bi-directional Predictive Coding Networks (DBPCNs). The synapses in a DBPCN allow both feedforward and feedback propagation of information using the same weights. This is in contrast to existing studies on PC which only allow feedback propagation to transmit predictions and the errors in these predictions are used to update representations and weights \cite{rao1999predictive}. In DBPCN, each layer simultaneously predicts the activities of neurons in both, previous layer using feedback propagation and next layer using feedforward propagation. The errors in these predictions are used to estimate representations associated with each layer and the weights in the network. Once trained, feedforward propagation from the input to output layer is used for classification. Feedback propagation is used to reconstruct a given input based on representations associated with any given layer in a DBPCN. 

%In DBPCN, Feedforward propagation from the input to output layer is used for classification. Feedback propagation is used to reconstruct a given input based on representations associated with any given layer in DBPCN.

% DBPCN is trained using a novel learning algorithm termed Deep Bi-directional Predictive Coding (DBPC). DBPC extends PC \cite{rao1999predictive} by updating the representations in a given layer such that they can simultaneously predict the representations associated with the previous (feedback propagation) and next (feedforward propagation) layers. Similarly, the weights between any two layers of DBPCN are updated to minimize the errors in predictions generated using both, feedforward and feedback propagation using these weights. Note that the modifications introduced by DBPC still utilize only the locally available information for updating the representations and weights. 

The DBPC has been implemented in this paper for networks with both fully connected (DBPC-FCN) and convolutional layers (DBPC-CNN). The performance of these networks has been evaluated using MNIST and FashionMNIST datasets for both classification and reconstruction. The classification accuracy and the images reconstructed using DBPC-FCN and DBPC-CNN are similar to the existing best performing algorithms. For both types of problems, network trained using DBPC require fewer parameters and utilize local learning rules which support parallel learning across all layers in the network.
% The classification accuracy and the quality of reconstruction of DBPC have been reported to be comparable to the performance of earlier iterations of PC \cite{millidge2020relaxing}, variants of PC such as FIPC\cite{song2018fast} and other classical DNNs \cite{seng2021mnist} while using fewer parameters. Experimental results clearly indicate that DBPC can utilize locally available information to train the network and simultaneously achieve classification and reconstruction using the same weights. 
% DBPC also performs better than existing similar works while using fewer parameters.

% performance of DBPC is evaluated using MNIST and FashionMNIST datasets. The classification accuracy of convolutional DBPC (99.57\%) is better than FIPC\textsubscript{3} (98.64\%) \cite{song2018fast} and other classic DNNs \cite{seng2021mnist} by using fewer parameters. Meanwhile, the original input image can be reconstructed from all representation layers in DBPC by sharing the weights of the classification task.

The rest of the paper is organized as follows. Section \ref{section2} summarizes other approaches in literature that simultaneously perform classification and reconstruction. The architecture of DBPCN and its learning algorithm are presented in Section \ref{section3}. Experimental results using DBPCN are presented in Section \ref{section4}. Finally, Section \ref{section5} summarizes the conclusions from this study and identifies directions for the future.

\section{Related work}
\label{section2}
% Both error-backpropagation and PC have been utilized to develop networks that can simultaneously perform classification and reconstruction. Below, the review has been organized according to the learning algorithm used in the related works.

%\textit{Error-backpropagation} (EBP): Most existing methods that utilize EBP either perform classification or reconstruction. Recently, a novel classification supervised AE (CSAE) \cite{zhu2019classification} based on predefined evenly-distributed class centroids (PEDCC) is proposed. This model can realize competitive classification accuracy, encoding and decoding capabilities at the same time.

Over the last few decades, PC has emerged as an important theory of information processing in the brain \cite{mumford1992computational, friston2005theory}. Due to a lack of supervisory signal in the brain, most computational studies involving PC in neuroscience develop generative models using unsupervised forms of learning \cite{spratling2008predictive, spratling2008reconciling}. These studies clamp the activity associated with the input layer while neural activity in other layers is updated to estimate suitable representations. The layer representations estimated using these methods can be used to reconstruct the original input. In \cite{ororbia2022convolutional}, PC is further developed and used to train convolutional neural networks for image denoising on Color-MNIST and CIFAR-10.

%Figure1 Basic_PC_Learning
% \begin{figure}[!t]
% \centering
% \includegraphics[width=3.0in]{Images/Basic_PC_Learning.jpg}
% \caption{Training network of predictive coding. This network consists of feedback and feedforward connection, as shown the red and black arrows respectively. Dashed boxes represent the predictions about the activity of neurons in the previous layer and $\mathbf{e}_l$ denotes the prediction errors between generated prediction and the previous layer. The role of feedback propagation is to use local weights $\mathbf{W}_{l}$ and the activity of neurons $\mathbf{y}_{l+1}$ in the top layer of the network to predict the activity of neurons $\hat{\mathbf{y}}_l$ in down layer ($\hat{\mathbf{y}}_l=f(\mathbf{W}_{l}\mathbf{y}_{l+1})$). Feedforward connection is responsible to calculate the local error $\mathbf{e}_{l}$ between prediction $\hat{\mathbf{y}}_l$ and the down layer $\mathbf{y}_{l}$. Finally, the PC model can use locally available information, such as the top-down error $\mathbf{e}_{l}$ and bottom-up error $\mathbf{e}_{l-1}$, to infer the activity of neurons $\mathbf{y}_{l}$ and update weights $\mathbf{W}_{l}$.} 
% \label{fig1}
% \end{figure}

Several recent studies have developed supervised forms of PC \cite{sun2020predictive, millidge2020relaxing, song2018fast}. The key idea in these studies is to clamp activities associated with both, input and output layers to samples and corresponding labels, respectively during training. For testing, only the activity associated with the input layer is clamped to a given sample and the estimated output layer representations are utilized to predict class labels. The networks developed in the above-mentioned studies are only suitable for classification. In \cite{song2018fast}, PC is used to develop a network that simultaneously performs classification and reconstruction. However, this paper utilizes separate set of parameters in each layer for classification and reconstruction which increases its computational requirements. % Further, the classification accuracies reported in these studies are about 98\%, which is lower than the performance of DNNs (more than 99.4\%) trained using EBP.

% However, papers \cite{sun2020predictive} and \cite{millidge2020relaxing} can not reconstruct input samples when they achieved classification. 

In \cite{wen2018deep}, PC is used to develop networks that can simultaneously perform classification and reconstruction using the same set of parameters. PC is only used to estimate the representations associated with each layer. The weights in the network are updated using EBP which relies on non-local information to update the weights and is unsuitable for parallel learning across all layers.

This paper aims to develop a new method to develop networks that can simultaneously perform classification and reconstruction using the same connections. The representations and weights in the proposed method are learned using locally available information to support parallelization of learning across the network.

% Further, paper \cite{wen2018deep} proposed a deep predictive coding architecture (PCN), which adopts PC principle to improve CNNs architecture and uses EBP to update parameters. Compared to the state-of-the-art models, this method has achieved competitive classification accuracy and reconstruction quality. However, both above models use EBP to update parameters and cannot use locally available information to update parameters, which will lead to more expensive to realize massively parallel computation in hardware. Further, PCN requires several cycles of recursive computation, which will cost 2t times the FLOPs of the plain CNN \cite{wen2018deep}. 

\section{Deep Bi-directional Predictive Coding (DBPC)}
\label{section3}
DBPC can be used for networks with fully connected layers and convolutional neural networks. Here we describe 1) the computations for bi-directional propagation of information in DBPC using a Fully Connected Network (DBPC-FCN); 2) the learning algorithm for estimating representations and updating the weights in DBPC-FCN; and 3) a network architecture for using DBPC to train convolutional neural networks (DBPC-CNN).

\subsection{Network Architecture}
\label{section3.1}
\begin{figure}[!htb]
\centering
\includegraphics[width=3.0in]{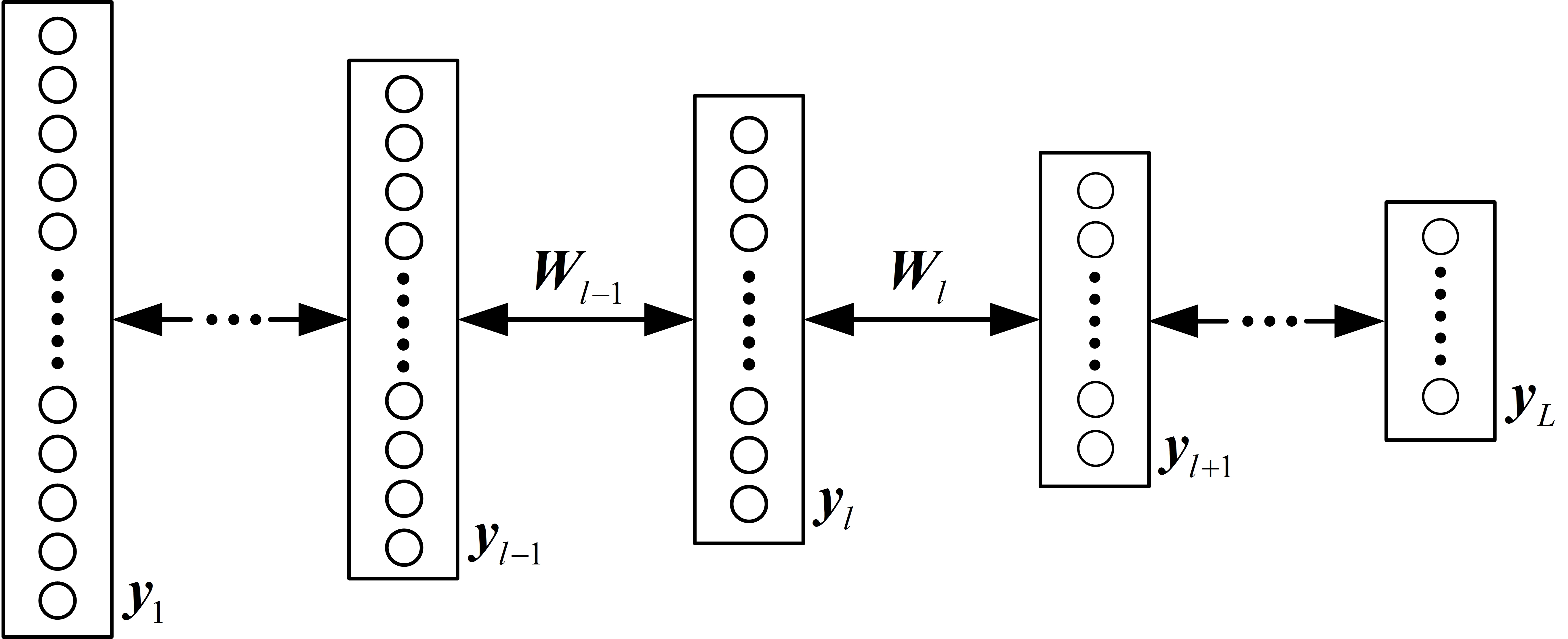}
\caption{Network architecture of DBPC-FCN with $L$ layers.}
\label{The architecture of DBPC-FCN}
\end{figure}

Fig. \ref{The architecture of DBPC-FCN} shows the architecture of the Deep Bi-directional Predictive Coding Fully Connected Network (DBPC-FCN) with $L$ layers. $\mathbf{y}_l$ is a vector of shape ($n_l\times1$) which represents the activity of neurons in the $l^{th}$ layer of the network. $n_l$ denotes the number of neurons in $l^{th}$ layer. DBPC-FCN employs bi-directional connections (black lines with arrows at both ends) between all layers of the network which enables information to propagate in both, feedforward and feedback directions.

%Basic_PC_Learning

Based on feedforward propagation from $(l-1)^{th}$ to $l^{th}$ layer, the activity of neurons in the $l^{th}$ layer is given by

\begin{equation}
\label{y_l_ff}
\hat{\mathbf{y}}_l^{ff}=f(\mathbf{W}_{l-1}\mathbf{y}_{l-1})
\end{equation}

where $f$ denotes the activation function and $\mathbf{W}_{l-1}$ is a ($n_l\times n_{l-1}$)  matrix which denotes the weights of the connections between $(l-1)^{th}$ and $l^{th}$ layer of the network. The Rectified Linear Unit (ReLU) is used as the activation function for all networks in this paper.

Similarly, when feedback propagation is used, the activity of neurons in the $l^{th}$ layer is determined using $(l+1)^{th}$ layer, given by

\begin{equation}
\label{y_l_fb}
\hat{\mathbf{y}}_l^{fb}=f(\mathbf{W}_l^T\mathbf{y}_{l+1})
\end{equation}
where $\mathbf{W}_l^T$ denotes the transpose of weights $\mathbf{W}_l$. Equations (\ref{y_l_ff}) and (\ref{y_l_fb}) represent the \emph{predictions} about the activity of neurons in the $l^{th}$ layer based on feedforward propagation from $(l-1)^{th}$ and $(l+1)^{th}$ layer, respectively (see Section \ref{section3}.$B$ for further explanation).  It should be noted that both, feedforward and feedback propagation employ the same set of weights.

While training the network an input sample is processed using both, feedforward and feedback propagation. During inference, feedforward propagation is utilized for classification tasks and feedback propagation is used to infer representations that allow reconstructing a given input. For classification, an input sample is presented through the first layer and the predicted class is determined by propagating information from the first layer to the $L^{th}$ layer. Given an input sample $\left(\mathbf{x}_k\right)$ and the associated class label $\left(\mathbf{c}_k\right)$, the goal of the learning algorithm is to estimate output layer representations that enable correct classification.

\subsection{Learning Algorithm}
\label{section3.2}
The goal of the DBPC is to estimate representations in all layers that can simultaneously be used for classification and reconstruction. The learning algorithm relies only on the locally available information to simultaneously infer representations and update weights in the network. For the $l^{th}$ layer in the network, locally available information includes activities of neurons in the previous $\left((l-1)^{th}\right)$ and next $\left((l+1)^{th}\right)$ layer, and the weights ($\mathbf{W}_{l-1}$  and $\mathbf{W}_l$ ) of the connections between these layers. The fundamental concept underlying the learning algorithm is that each layer in the network aims to predict the activities of neurons in the previous (feedback propagation) and next layer (feedforward propagation). The errors in these predictions form the basis of inferring suitable representations (representation learning) and updating the weights (model learning). Below, the two steps of the learning algorithm namely, representation learning and model learning are described in detail.

% DBPC uses the proposed learning algorithm to achieve end-to-end training. The input samples and label pairs \{$\mathbf{x}_k$, $\mathbf{c}_k$\} of the training set are clamped to the first layer ($\mathbf{y}_1=\mathbf{x}_k$) and the last layer ($\mathbf{y}_{l=max}=\mathbf{c}_k$) during training, respectively. This clamping strategy provides the network with a source of information for reconstruction and classification. After multiple iterations of updating each representation according to below representation learning, the last layer ($\mathbf{y}_{l=max}$) will gradually affect each representation layer. It may be noted that an input sample can be reconstructed using representations inferred by DBPC in any layer of the network. Given representations associated with a particular layer, information is sequentially propagated backwards till the input layer to obtain a reconstruction of the input.

% \subsubsection*{\bf Representation learning}
\subsubsection{Representation learning}
\label{section3.2.1}
Using feedforward propagation, $l^{th}$ layer in the network receives a prediction of its own neuronal activity from the $(l-1)^{th}$ layer (Equation (\ref{y_l_ff})) and generates a prediction about the activities of neurons in the $(l+1)^{th}$ layer. Based on feedforward propagation, the error $\left(\mathbf{e}_{l-1}^{ff}\right)$ in the prediction about the activity of neurons in the $l^{th}$ layer is given by

\begin{equation}
\label{e_l_ff}
\mathbf{e}_{l-1}^{ff}=(\mathbf{y}_{l}-\hat{\mathbf{y}}_l^{ff})^2
\end{equation}

% Since feedforward propagation uses the previous layer $\mathbf{y}_{l}$ to predict the next layer $\mathbf{y}_{l+1}$ and the input samples and labels are clamped to the first layer and the last layer, minimizing the above errors $\mathbf{e}_{l-1}^{ff}$ and $\mathbf{e}_{l}^{ff}$ will gradually make the generated representations map to the real labels after multiple iterations of updating each representation.

Similarly, using feedback propagation, $l^{th}$ layer in the network receives a prediction (Equation (\ref{y_l_fb})) of its own activities from the $(l+1)^{th}$ layer and generates a prediction about the activities of neurons in the $(l-1)^{th}$ layer. Based on feedback propagation, the error $\left(\mathbf{e}_{l}^{fb}\right)$ in the prediction about the activity of neurons in the $l^{th}$ layer is given by

\begin{equation}
\label{e_l_fb}
\mathbf{e}_{l}^{fb}=(\mathbf{y}_{l}-\hat{\mathbf{y}}_{l}^{fb})^2
\end{equation}
%On the contrary, as feedback propagation uses the last layer $\mathbf{y}_{l+1}$ to predict the previous layer $\mathbf{y}_{l}$, minimizing the above errors $\mathbf{e}_{l}^{fb}$ and $\mathbf{e}_{l+1}^{fb}$ will gradually make the generated representations map to the input samples after multiple iterations of updating each representation.

\begin{figure}[!htb]
\centering
\includegraphics[width=3.0in]{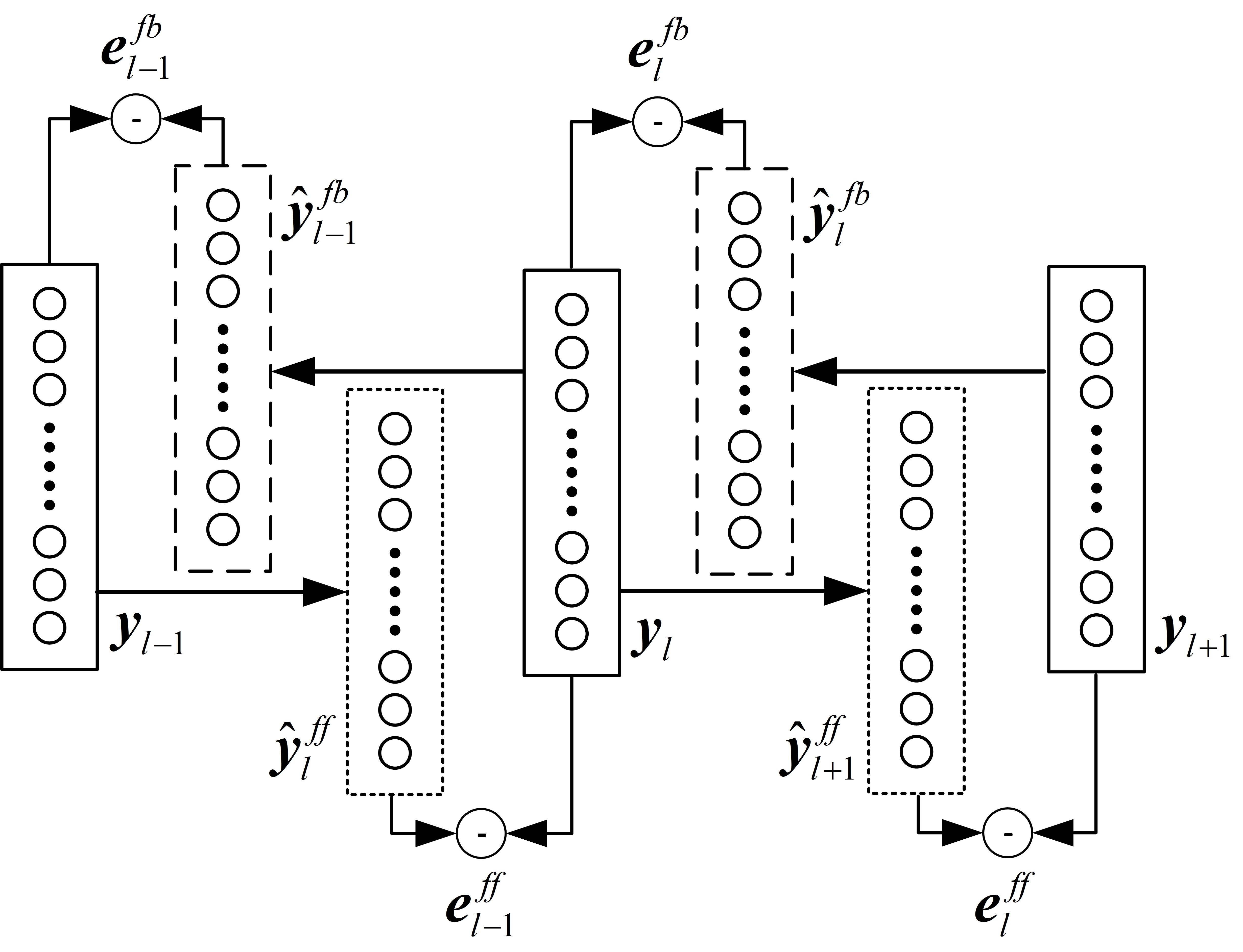}
\caption{Visualization of the locally computed errors for representation learning and model learning in DBPC. The dotted and dashed rectangles represent the feedforward and feedback predictions, respectively. The circles represent item-wise subtraction required to compute errors in feedforward and feedback predictions.}
\label{The training network of DBPC}
\end{figure}

Figure \ref{The training network of DBPC} shows a visualization for computation of all locally computed errors that involve representations $\left(\mathbf{y}_l\right)$ associated with $l^{th}$ layer in the network. $\mathbf{y}_l$ is updated by performing gradient descent on all the locally computed errors, given by
\begin{equation}
\label{E_y_l}
\mathbf{E}_{y_l}=\lambda_f(\mathbf{e}_{l-1}^{ff}+\mathbf{e}_{l}^{ff})+\lambda_b(\mathbf{e}_{l-1}^{fb}+\mathbf{e}_{l}^{fb})
\end{equation}
where $\lambda_f$ and $\lambda_b$ denote feedforward and feedback factors, respectively. $\lambda_f$ controls the impact of errors in feedforward predictions on the updated representations. Similarly, $\lambda_b$ determines the influence of errors in feedback predictions on the updated representations.

%Figure Fully_PC_Learning

Minimizing the errors in feedback predictions improves the reconstructions generated by the network and reducing the errors in feedforward predictions improves the classification accuracy of the network. Thus, suitable values for $\lambda_f$ and $\lambda_b$ help the network to simultaneously perform well on classification and reconstruction tasks. 

Based on the error in equation (\ref{E_y_l}), the update $\left(\Delta\mathbf{y}_{l}\right)$ in the representations associated with $l^{th}$ layer are given by
\begin{equation}
\label{Delta_y_l}
\Delta\mathbf{y}_{l}=-\ell_y\frac{\delta\mathbf{E}_{y_l}}{\delta\mathbf{y}_l}
\end{equation}
where $\ell_y$ denotes the learning rate for updating representation. The representations are updated using the Equations (\ref{y_l_ff})-(\ref{Delta_y_l}) multiple times as in the original PC algorithm \cite{rao1999predictive}. In this paper, the representations are updated 20 times in DBPC-FCN. Since, all the information required to compute the error in Equations (\ref{E_y_l}) is available locally, the representations for all layers are updated in parallel.

\subsubsection{Model Learning}
\label{section3.2.2}
The weights in DBPCN are also updated using only locally available information. The weights between $(l-1)^{th}$ and $l^{th}$ layers of the network are updated to minimize the errors in predictions based on feedforward and feedback propagation involving $\mathbf{W}_{l}$. Thus, $\mathbf{W}_{l}$ is updated by performing gradient descent on the errors in Equation (\ref{e_l_ff}) and (\ref{e_l_fb}), given by

\begin{equation}
\label{E_W_l}
\mathbf{E}_{W_l}=\beta_c\mathbf{e}_{l}^{ff}+\beta_r\mathbf{e}_{l}^{fb}
\end{equation}

where $\beta_c$ and $\beta_r$ denote classification and reconstruction factors for updating weights, respectively. $\beta_c$ controls the change in weight to improve the feedforward predictions and hence, the classification performance of the DBPCN. Similarly, $\beta_r$ determines the change in weight to improve the feedback predictions and hence, the reconstruction performance of the network.
%Similarly, estimating $\mathbf{W}_{l}$  by minimizing the errors in Equations (\ref{equation3}) improves the classification performance of the network, and minimizing the errors in Equations (\ref{equation4}) improves the reconstructions obtained from the network. 
Suitable values for $\beta_c$ and $\beta_r$ enable the network to simultaneously achieve good performance on classification and reconstruction tasks. 

Based on the error in equation (\ref{E_W_l}), the change in $\mathbf{W}_{l}$ is given by
\begin{equation}
\label{Delta_W_l}
\Delta\mathbf{W}_{l}=-\ell_w \frac{\delta\mathbf{E}_{W_l}}{\delta\mathbf{W}_l}
\end{equation}
where $\ell_w$ denotes the learning rate for updating weights. The locally computed error for updating weights ensures that all weights in the network can be updated in parallel. Algorithm \ref{alg:alg1} presents the pseudocode for the DBPC learning algorithm.

\begin{algorithm}[H]
    \caption{Learning algorithm for DBPC}\label{alg:alg1}
    \begin{algorithmic}[1]
    \Require{Samples and labels \{$\mathbf{x}_k$, $\mathbf{c}_k$\}}
      \For{each epoch}
          \For{each sample}
            \State{\textcolor{gray}{\% Clamp the first and last layer to $\mathbf{x}_k$ and $\mathbf{c}_k$}}
            \State{{$\mathbf{y}_{1}=\mathbf{x}_k$},\hspace{0.2cm}{$\mathbf{y}_{L}=\mathbf{c}_k$}}
                \State{\textcolor{gray}{\% Feedforward propagation}}
                % \State{${\mathbf{y}}_l=f(\mathbf{W}_{l-1}\mathbf{y}_{l-1})$}
                \State{$\hat{\mathbf{y}}_l^{ff}=f(\mathbf{W}_{l-1}\mathbf{y}_{l-1})$}
                \State{\textcolor{gray}{\% Feedback propagation}}
                \State{$\hat{\mathbf{y}}_l^{fb}=f(\mathbf{W}_l^T\mathbf{y}_{l+1})$}            
            \For{each iteration}
                \State{\textcolor{gray}{\% Compute errors in Equation \ref{e_l_ff} and \ref{e_l_fb}}}
                \State{$\mathbf{e}_{l-1}^{ff}=(\mathbf{y}_{l}-\hat{\mathbf{y}}_l^{ff})^2$}
                \State{$\mathbf{e}_{l}^{fb}=(\mathbf{y}_{l}-\hat{\mathbf{y}}_{l}^{fb})^2$}
                \State{\textcolor{gray}{\% Update representation}}

                \State{$\Delta\mathbf{y}_{l}=-\ell_y\frac{\delta\mathbf{E}_{y_l}}{\delta\mathbf{y}_l}$}, $\forall l \in [2, \cdots, (L-1)]$   
            \EndFor
            \State{\textcolor{gray}{\%Update weights}}
            \State{$\Delta\mathbf{W}_{l}=-\ell_w \frac{\delta\mathbf{E}_{W_l}}{\delta\mathbf{W}_l}$}, $\forall l \in [1, \cdots, (L-1)]$             
          \EndFor
      \EndFor
    \end{algorithmic}
\end{algorithm}

During testing, only the activities of the neurons in the first layer are clamped to a given input. The activities of neurons in all the other layers of the network are estimated using representation learning. The predicted class is estimated based on the representations associated with the output layer neurons. Further, estimated representations for any other layer can be used to reconstruct the given input using feedback propagation.

% After training, the weights $\mathbf{W}_{l}$ are used for testing. Since all errors do not need to be calculated to update representations and weights during testing, the feedforward propagation only needs to be run once to infer each representation layer and obtain the classification. Then, the input samples are reconstructed through feedback propagation using each representation layer inferred by feedforward propagation. Algorithm \ref{algorithm2} presents the pseudocode for DBPC testing algorithm.

% The training network of DBPC is shown in Fig. \ref{The training network of DBPC}, which presents the generation of predictions and errors according to the locally available information. Compared with the architecture of DBPC-FCN, the training network splits the black line with arrows at both ends into two black lines with single arrow. Based on feedforward propagation from $(l-1)^{th}$ to $l^{th}$ layer, the prediction $\hat{\mathbf{y}}_l^{ff}$ about the activity of neurons in the $l^{th}$ layer is represented by dot-dash box. Similarly, Dashed box represents the predictions $\hat{\mathbf{y}}_l^{fb}$ about the activity of neurons in the $l^{th}$ layer based on feedback propagation from $(l+1)^{th}$ to $l^{th}$ layer. The training network of DBPC also adds black arrows to present the errors $\mathbf{e}_l^{ff}$ based on feedforward propagation and errors $\mathbf{e}_l^{fb}$ based on feedback propagation. 

\subsection{DBPC for Convolutional Neural Networks (DBPC-CNN)}
\label{section3.3}
We have also developed a network architecture to use DBPC for Convolutional Neural Networks (DBPC-CNN). To enable feedforward and feedback propagation using the same kernels in DBPC-CNN, each layer employs a padding $(P)$, given by
\begin{equation}
    P = \frac{K - 1}{2}
\end{equation}
where $K$ is the size of the kernel and stride in all layers is set to 1. Choosing padding in this way ensures that both input and output for a convolution operation have the same shape. This ensures that convolution operation can be applied in both feedforward and feedback direction using the same kernel.

\section{Experiments}
\label{section4}
This section presents the results of performance evaluation of DBPC for classification and reconstruction tasks. The performance of DBPC is also compared with the other existing algorithms for both tasks. The classification accuracy of DBPC is compared with other PC approaches, namely FIPC\textsubscript{3} \cite{song2018fast}, PC-1\cite{millidge2020relaxing} and PCN-E-1 \cite{wen2018deep}. In addition, the classification accuracy is also compared with the performance of classical DNNs, which include MobileNet-v2 \cite{seng2021mnist} and GoogLeNet \cite{seng2021mnist}. The reconstruction performance of DBPC is compared with the performance of FIPC\textsubscript{3} which is the only other PC approach that is simultaneously capable of classification and reconstruction using representations from any layer in the network.

The performance of DBPC is evaluated in terms of the number of network parameters  and accuracy for classification.  Given a confusion matrix $Q$, the classification accuracy ($\eta_c$) is given by
\begin{equation}
\label{acc_eq}
\eta_c=\frac{\Sigma_{i \in \{1, \cdots, N_C\}} c_{ii}}{\Sigma_{i,j \in \{1, \cdots, N_C\}} c_{ij}}
\end{equation}
where $c_{ij}$ represents the values in $i^{th}$ row and $j^{th}$ column of the confusion matrix and $N_C$ denotes the total  number of classes. The Peak Signal-to-Noise Ratio (PSNR) and Structural Similarity Index Measure (SSIM) are used for comparing performance on reconstruction task. The PSNR \cite{hore2010image} $(\eta_r)$ is given by
\begin{equation}
\label{psnr}
\eta_r=10\times\log_{10}\frac{MAX^2}{MSE}
\end{equation}
where $MAX$ represents the maximum pixel intensity in the image and $MSE$ is the mean squared error between the original image and the reconstructed image. The SSIM \cite{wang2004image} $(\eta_s)$ is given by
\begin{equation}
\label{ssim}
\eta_s(x,y)=\frac{(2\mu_x\mu_y+C_1)(2\sigma_{xy}+C_2)}{(\mu_x^2+\mu_y^2+C_1)(\sigma_x^2+\sigma_y^2+C_2)}
\end{equation}
where $x$ and $y$ are the original and the reconstructed images, respectively. $\mu_x$ and $\mu_y$ represent mean pixel intensity of $x$ and $y$, respectively. $\sigma_x$ and $\sigma_y$ represent the standard deviations of the pixel intensities in $x$ and $y$, respectively. $\sigma_{xy}$ denotes the covariance of pixel intensities across $x$ and $y$. $C_1$ and $C_2$ are constants to prevent division by zero.

The performance evaluation is conducted using the MNIST \cite{lecun1998gradient} and FashionMNIST \cite{xiao2017fashion} datasets. MNIST is a dataset that contains images of hand-written digits from 0 to 9. It has 60,000 grayscale images for training and 10,000 grayscale images for testing. The FashionMNIST dataset is a more challenging dataset that contains images of ten fashion items like T-shirts, trousers and bags. Similar to MNIST, FashionMNIST contains 60,000 grayscale images for training and 10,000 grayscale images for testing. Each image in both datasets is of the size $28\times28$ pixels.

Table \ref{Architecture for DBPC} shows the architectures for DBPC-FCN and DBPC-CNN used for the two datasets. A given row in the table shows details of the corresponding layer in the network. The performance on the MNIST dataset has been evaluated using a fully connected network and a convolutional neural network. For the FashionMNIST dataset, the performance of DBPC is evaluated using only a convolutional neural network. The number of neurons in each layer of DBPC-FCN has been shown using the prefix `FC'. Similarly, the prefix `Conv' is used to specify the number of channels in a particular convolutional layer of DBPC-CNN. All convolutional layers use kernel, padding and stride of 3, 1 and 1, respectively.

%The first one is fully connected DBPCN. It contains 3 different the number of fully connected layers, named A and B respectively. The numbers in each row in the table represent the number of nodes in each layer. Another one is convolutional DBPC-CNN. It also contains 3 different the number of convolutional layers, named C and D respectively. The numbers in Conv-x indicate the number of kernels in each convolutional layer. To make feedforward and feedback propagation use the same set of weights, padding and stride operations are performed in each convolutional layer. Their values are 3 and 1, respectively.

\begin{table}[!htb]
\centering
\caption{\text{Architecture for DBPC-FCN and DBPC-CNN}}
\begin{tabular}{c|ccc}
\hline
                  
                  {Dataset}& \multicolumn{2}{c|}{MNIST}  & \multicolumn{1}{c}{FashionMNIST}                            \\ \hline
                  {Architecture}& \multicolumn{1}{c|}{DBPC-FCN} & \multicolumn{1}{c|}{DBPCN-CNN} & {DBPC-CNN}                            \\ \hline
                  %{Architecture}& \multicolumn{1}{c|}{A} & \multicolumn{1}{c|}{B} & \multicolumn{1}{c|}{C} & {D} \\ \hline
                  %{\#Layers}& \multicolumn{1}{c|}{4 layers} & \multicolumn{1}{c|}{6 layers} & {10 layers} \\ \hline
                  {Input Size}& \multicolumn{3}{c}{$(28\times28)$}      \\ \hline
                  \multirow{9}{*}{\begin{tabular}{@{}c@{}}Number\\of\\neurons\\in\\a\\layer\end{tabular}} & \multicolumn{1}{c|}{FC-1000}   & \multicolumn{1}{c|}{Conv-16} & {Conv-16} \\ \cline{2-4} 
                  & \multicolumn{1}{c|}{FC-400} & \multicolumn{1}{c|}{Conv-32} & {Conv-32}  \\ \cline{2-4} 
                  & \multicolumn{1}{c|}{FC-100} & \multicolumn{1}{c|}{Conv-32} & {Conv-32} \\ \cline{2-4} 
                  & \multicolumn{1}{c|}{} & \multicolumn{1}{c|}{Conv-48} & {Conv-48} \\ \cline{2-4} 
                  & \multicolumn{1}{c|}{} & \multicolumn{1}{c|}{Conv-48} & {Conv-48} \\ \cline{2-4} 
                  & \multicolumn{1}{c|}{} & \multicolumn{1}{c|}{} & {Conv-64} \\ \cline{2-4} 
                  & \multicolumn{1}{c|}{} & \multicolumn{1}{c|}{} & {Conv-64} \\ \cline{2-4} 
                  & \multicolumn{1}{c|}{} & \multicolumn{1}{c|}{} & {Conv-96} \\ \cline{2-4} 
                  & \multicolumn{1}{c|}{} & \multicolumn{1}{c|}{} &  {Conv-96} \\ \hline
                  {Classification} & \multicolumn{3}{c}{FC-10}  \\ \hline
                  {\#Parameters} & \multicolumn{1}{c|}{1.225M}  & \multicolumn{1}{c|}{0.425M} &{1.004M} \\ \hline
\end{tabular}
\label{Architecture for DBPC}
\end{table}

All models presented in this paper have been implemented in PyTorch and trained using an Nvidia V100 GPU. Training data is augmented using random rotation and affine transformations. A minibatch of 32 is used during training and stochastic gradient descent (SGD) is used to optimize the network parameters. The total number of epochs is set to 50 and 100 on MNIST and FashionMNIST datasets, respectively.

\subsection{Performance Comparison for Classification}
\label{section4.1}
\begin{figure}[!htb]
\centering
\includegraphics[width=3.5in]{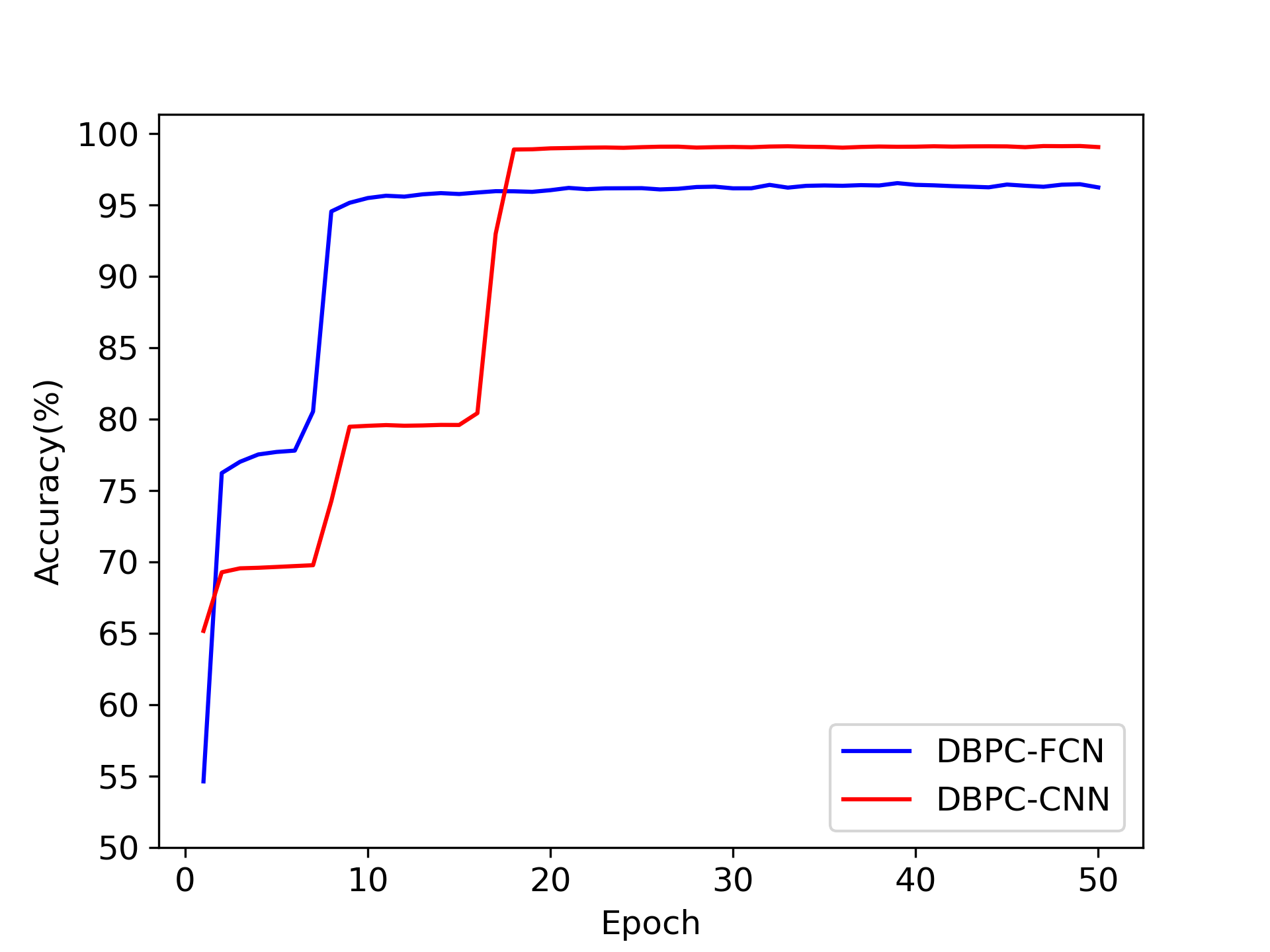}
\caption{Classification accuracy of DBPC-FCN and DBPC-CNN on MNIST after each epoch of training.}
\label{Training classification accuracy on MNIST}
\end{figure}

Table \ref{Performance comparison on MNIST} shows the results of performance comparison between DBPC and other existing learning algorithms for classification on the MNIST dataset. Figure \ref{Training classification accuracy on MNIST} shows how the classification accuracy of DBPC-FCN and DBPC-CNN evolves during training for the MNIST dataset.

\begin{table}[!htb]
\centering
\caption{Performance comparison of DBPC with other methods on the MNIST dataset}
\begin{tabular}{c|c|c}
\hline
\multicolumn{1}{c|}{Methods} & \multicolumn{1}{c|}{Testing Accuracy {$\eta_c$} (\%)} & \multicolumn{1}{c}{Parameters} \\ \hline
\multicolumn{3}{c}{Fully Connected Networks} \\ \hline
                    {FIPC\textsubscript{3}\cite{song2018fast}}           &  {98.84}   &   {2.450M}         \\ \hline
                    {PC-1\cite{millidge2020relaxing}}           &  {98.00}   &   {0.532M}         \\ \hline
                    {PC-2\cite{sun2020predictive}}          &  {98.00}   &   {1.672M}         \\ \hline
                    {DBPC-FCN (Proposed work)}          &  {97.67}   &    {1.225M}        \\ \hline
\multicolumn{3}{c}{Convolutional Neural Networks} \\ \hline
                    {PCN-E-1(tied) \cite{wen2018deep}}  &  {99.57}   &    {0.070M}        \\ \hline
                    %{CSAE\cite{zhu2019classification}}             &  {99.52}   &    {-}             \\ \hline
                    {MobileNet-v2\cite{seng2021mnist}}    &  {99.43}   &    {13.600M}         \\ \hline
                    {GoogLeNet\cite{seng2021mnist}}       &  {99.47}   &    {49.700M}         \\ \hline
                    
                    % {DBPCN\_B}           &  {97.64}   &    {2.015M}        \\ \hline
                    {DBPC-CNN (Proposed work)}       &  {99.33}   &    {0.425M}        \\ \hline
                    % {DBPC-CNN\_D }       &  {\bf{99.57}}   &    {1.004M}        \\ \hline 
\end{tabular}
\label{Performance comparison on MNIST}
\end{table}

%ACCURACY

DBPC-FCN uses a network with 1.225 million parameters to achieve a classification accuracy of 97.67\% which is 1.2\% lower than the best-performing method. FIPC\textsubscript{3} is the best-performing algorithm with an accuracy of 98.84\% but it uses twice the number of parameters used by DBPC-FCN. PC-1 uses the smallest network with 0.532 million parameters to achieve an accuracy of 98.00\%. It may be noted that representations estimated in both, PC-1 and PC-2 can't be used for reconstruction whereas DBPC-FCN also supports reconstruction of inputs.

% DBPC-FCN achieved an accuracy of 97.67\% which is similar to the classification accuracy of other fully connected PC networks \cite{song2018fast, millidge2020relaxing, sun2020predictive} using 1.225 million parameters. However, DBPC-CNN achieved significantly higher classification accuracy (99.33\%) than DBPC-FCN using less parameters (0.425 million).  The testing classification accuracy of DBPC-FCN and DBPC-CNN with epochs growth is shown in Fig. \ref{fig3}. From the above experimental results, DBPC-CNN can use fewer parameters to obtain higher classification accuracy than fully connected DBPC-FCN. 

The performance of all the methods used for comparison is better using convolutional neural networks. DBPC-CNN employs a network with 0.425 million parameters to achieve an accuracy of 99.33\% which is similar to the performance of other learning algorithms used for comparison. PCN-E-1 (tied) is the best performing algorithm with an accuracy of 99.57\% and it uses a network with 0.07 million parameters. It may be noted that PCN-E-1 uses error-backpropagation for training which relies on non-local information for learning and is not suitable for parallel training across layers in the network. DBPC-CNN allows reconstruction of inputs using representations estimated for any layer in the network. The ability of PCN-E-1 to reconstruct images using representations estimated for different layers hasn't been studied. The performance of DBPC-CNN is also similar to the classification accuracy of established networks like MobileNet-v2 \cite{seng2021mnist} and GoogLeNet \cite{seng2021mnist} which are not capable of reconstruction. Further, DBPC-CNN employs a network that is much smaller than those used by MobileNet-v2 and GoogLeNet.

\begin{table}[!htb]
\centering
\caption{Performance comparison of DBPC-CNN with PC-1 on the FashionMNIST dataset}
\begin{tabular}{c|c|c}
\hline
\multicolumn{1}{c|}{Methods} & \multicolumn{1}{c|}{Testing Accuracy {$\eta_c$} (\%)} & \multicolumn{1}{c}{Parameters} \\ \hline
                    {PC-1\cite{millidge2020relaxing}}           &  {89.00}   &   {0.532M}         \\ \hline
                    %{CSAE\cite{zhu2019classification}}           &  {92.89}   &   {-}         \\ \hline                    
                    {DBPC-CNN }      &  {91.61}   &   {1.004M}        \\ \hline 
                    % {DBPC-CNN\_E }      &  {\bf{92.27}}   &   {38.869MM}        \\ \hline                     
\end{tabular}
\label{Performance comparison on FashionMNIST}
\end{table}

\begin{figure}[!t]
\centering
\includegraphics[width=3.5in]{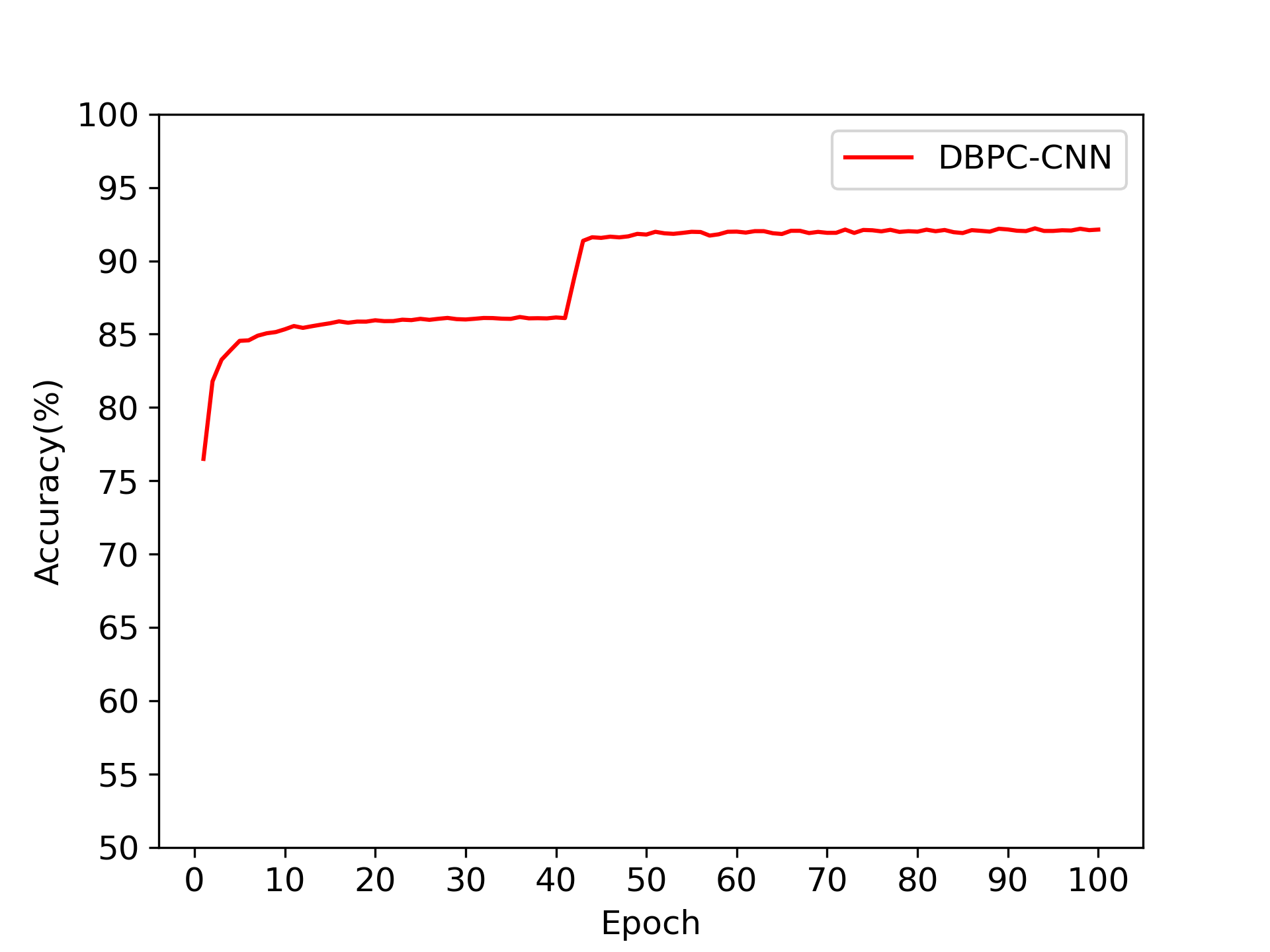}
\caption{Classification accuracy of DBPC-CNN on FashionMNIST after each epoch of training.}
\label{Training classification accuracy on FashionMNIST}
\end{figure}

Table \ref{Performance comparison on FashionMNIST} shows a performance comparison of DBPC-CNN with other methods on the more challenging FashionMNIST dataset. It may be noted that only DBPC-CNN is used for the FashionMNIST dataset due to the higher complexity of this dataset. Figure \ref{Training classification accuracy on FashionMNIST} shows the changes in the classification accuracy of DBPC-CNN after each epoch of training for the FashionMNIST dataset. DBPC-CNN uses a network with 1.004 million parameters to achieve an accuracy of 91.61\%. The performance of DBPC-CNN is 2.9\% higher than the classification accuracy of PC-1. Furthermore, as highlighted above, the representations estimated in PC-1 cannot be used for reconstructing the inputs.

\subsection{Performance Comparison for Reconstruction}
\label{section4.2}
\begin{figure}[!t]
	\centering 
        \begin{subfigure}[b]{0.5\textwidth}
            \includegraphics[width=\textwidth]{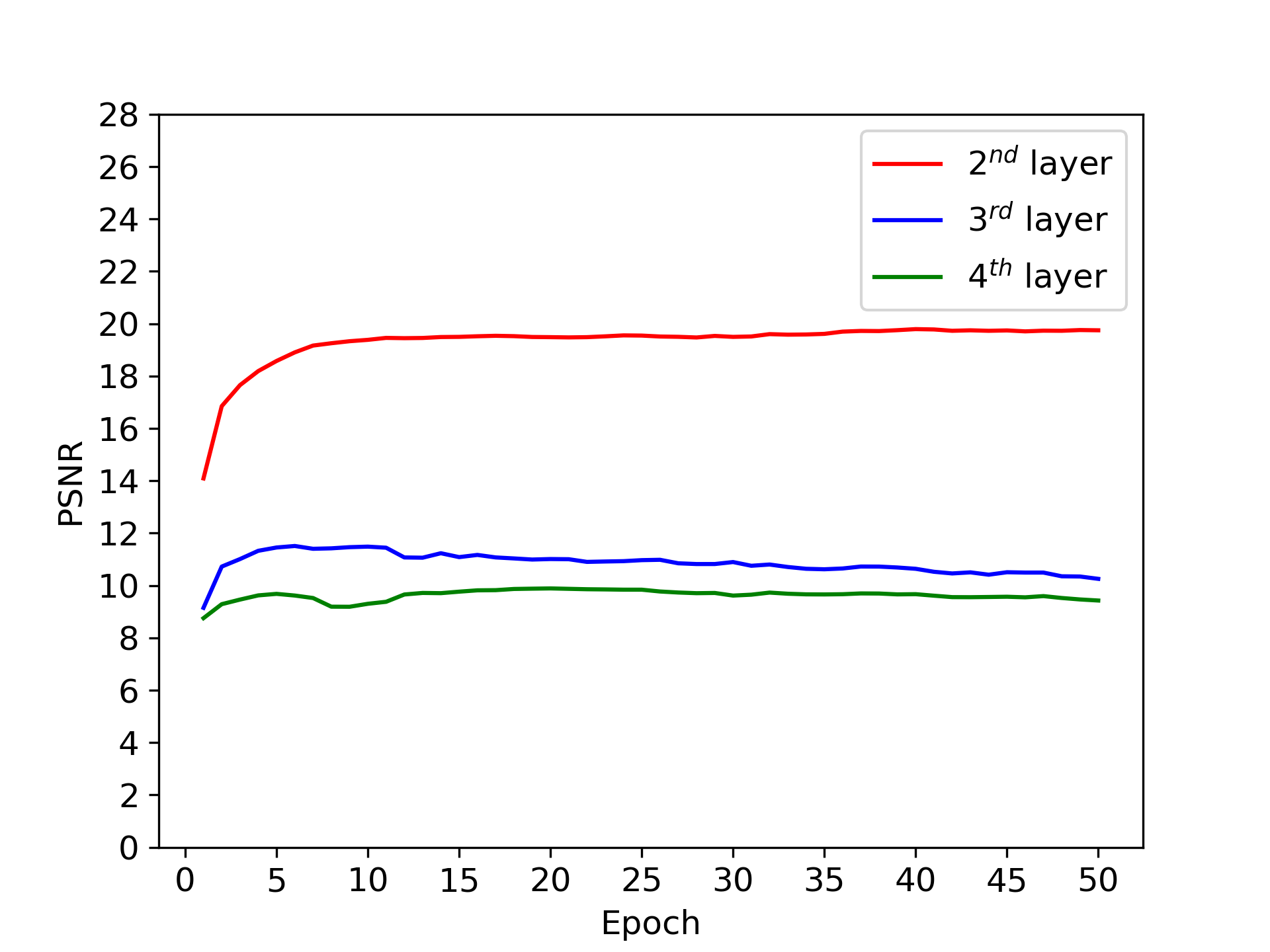}
            \caption{PSNR of DBPC-FCN}
            \label{The PSNR of DBPC-FCN}
        \end{subfigure}
        \begin{subfigure}[b]{0.5\textwidth}
            \includegraphics[width=\textwidth]{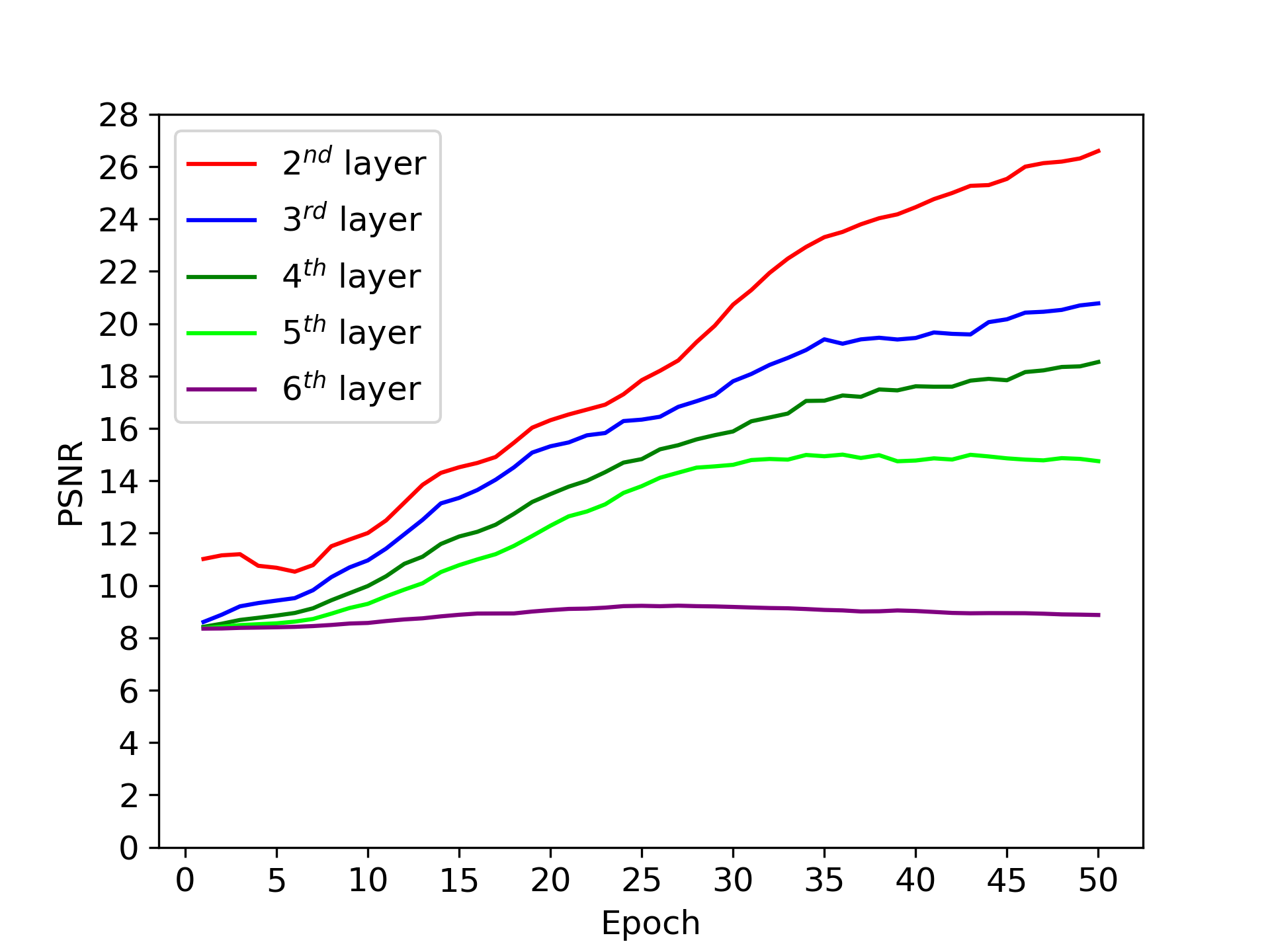}
            \caption{PSNR of DBPC-CNN}
            \label{The PSNR of DBPC-CNN}
        \end{subfigure}
	\caption{PSNR of (a) DBPC-FCN and (b) DBPC-CNN on MNIST after each epoch of training, respectively.}
	\label{The PSNR on MNIST}
\end{figure}

In this section, the performance of DBPC-FCN and DBPC-CNN is evaluated and compared for reconstruction problems using MNIST and FashionMNIST datasets. Figure \ref{The PSNR of DBPC-FCN} and \ref{The PSNR of DBPC-CNN} show how the PSNR of the reconstructed images form each layer in DBPC-FCN and DBPC-CNN evolves during training, respectively. For both DBPC-FCN and DBPC-CNN, earlier layers achieved a higher PSNR compared to deeper layers in the network. Further, reconstructed images obtained using DBPC-CNN exhibited higher PSNR compared to DBPC-FCN. Similar results are also obtained for the SSIM based on reconstructed images obtained from DBPC-FCN and DBPC-CNN. % Similarly, Figure \ref{The SSIM of DBPC-FCN} and \ref{The SSIM of DBPC-CNN} show how the SSIM of the reconstructed images changes with each epoch of training for DBPC-FCN and DBPC-CNN, respectively.

%PSNR

\begin{figure}[!t]
	\centering 
        \begin{subfigure}[b]{0.165\textwidth}
            \includegraphics[width=\textwidth]{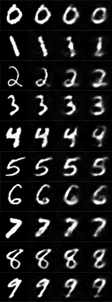}
            \caption{FIPC\textsubscript{3}}
            \label{FIPC_Recon}
        \end{subfigure}
        \begin{subfigure}[b]{0.18\textwidth}
            \includegraphics[width=\textwidth]{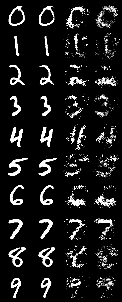}
            \caption{DBPC-FCN}
            \label{DBPC-A_MN}
        \end{subfigure}
        \begin{subfigure}[b]{0.28\textwidth}
            \includegraphics[width=\textwidth]{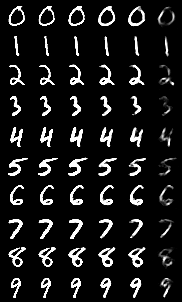}
            \caption{DBPC-CNN}
            \label{DBPC-D_MN}
        \end{subfigure}
	\caption{MNIST images reconstructed using (a) FIPC\textsubscript{3}, (b) DBPC-FCN and (c) DBPC-CNN, respectively. The first column in each panel represents the original image from the MNIST dataset, and the other columns from left to right display the reconstructions from a given layer in the network.}
\label{Reconstruction on MNIST}
\end{figure}

Figure \ref{Reconstruction on MNIST} shows the reconstructed images obtained using representations estimated for each layer in FIPC\textsubscript{3}, DBPC-FCN and DBPC-CNN. The first column in each figure shows the original figure in the dataset and the following columns show the reconstructions obtained from successively deeper layers in the network. These reconstructions are obtained by propagating backward from a given layer using the representations estimated in that layer. It may be observed that the quality of reconstructed images deteriorates as we go from earlier to deeper layers in all three algorithms. The deterioration in image quality is lowest for DBPC-CNN. It may be noted that FIPC\textsubscript{3} uses a separate set of weights for classification and reconstruction which results in a network having a large number of parameters.

%Reconstruction on MNIST

\begin{figure}[!t]
\centering
\includegraphics[width=3.0in]{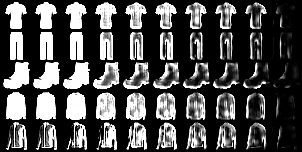}
\caption{Reconstruction results of DBPC-CNN on FashionMNIST. The first column in each subplot represents the original input image from the test FashMNIST dataset, and the other columns from left to right display the reconstructions from each representation layer.}
\label{Reconstruction on FashMNIST}
\end{figure}

% Figure \ref{PSNR on FashMNIST} and \ref{SSIM on FashMNIST} show how the PSNR and SSIM of the reconstructed images evolves with more epochs of training for DBPC-CNN on the more challenging FashionMNIST dataset, respectively. 
Figure \ref{Reconstruction on FashMNIST} show the images reconstructed using representations associated with each layer of DBPC-CNN for the FashionMNIST dataset. The layout of Figure \ref{Reconstruction on FashMNIST} and the method used for reconstructing these images is the same as that used for the MNIST dataset. Table \ref{Quantitative comparison using the PSNR and SSIM} provides a summary of the results presented above using a quantitative comparison of images reconstructed by DBPC-FCN and DBPC-CNN based on the PSNR and SSIM metric on both datasets.

%Reconstruction on FashMNIST

\begin{table}[!htb]
\centering
\caption{Quantitative comparison of images reconstructed by DBPC-FCN and DBPC-CNN based on PSNR and SSIM}
\begin{tabular}{l|l|l|l|l}
\hline
\multicolumn{1}{c|}{Methods} &  \multicolumn{1}{c|}{Datasets}   & \multicolumn{1}{c|}{$l^{th}$ layer} &  \multicolumn{1}{c|}{\begin{tabular}{@{}c@{}}Testing \\ PSNR ($\eta_r$)\end{tabular}} &  {\begin{tabular}{@{}c@{}}Testing \\ SSIM ($\eta_s$)\end{tabular}} \\ \hline
\multirow{3}{*}{DBPC-FCN} & \multirow{8}{*}{MNIST} & \multicolumn{1}{c|}{$2^{nd}$ layer} & \multicolumn{1}{c|}{20.21} & {0.9169} \\ \cline{3-5} 
                  &                   & \multicolumn{1}{c|}{$3^{rd}$ layer} & \multicolumn{1}{c|}{10.29} & {0.3851} \\ \cline{3-5} 
                  &                   & \multicolumn{1}{c|}{$4^{th}$ layer} & \multicolumn{1}{c|}{9.45}  & {0.2223} \\ \cline{1-1} \cline{3-5} 
\multirow{5}{*}{DBPC-CNN} &           & \multicolumn{1}{c|}{$2^{nd}$ layer} & \multicolumn{1}{c|}{26.90} & {0.9779} \\ \cline{3-5} 
                  &                   & \multicolumn{1}{c|}{$3^{rd}$ layer} & \multicolumn{1}{c|}{20.58} & {0.9228} \\ \cline{3-5} 
                  &                   & \multicolumn{1}{c|}{$4^{th}$ layer} & \multicolumn{1}{c|}{19.17} & {0.8967} \\ \cline{3-5} 
                  &                   & \multicolumn{1}{c|}{$5^{th}$ layer} & \multicolumn{1}{c|}{17.05} & {0.8304} \\ \cline{3-5} 
                  &                   & \multicolumn{1}{c|}{$6^{th}$ layer} & \multicolumn{1}{c|}{10.88} & {0.3832} \\ \hline
\multirow{9}{*}{DBPC-CNN} & \multirow{9}{*}{FashionMNIST} & \multicolumn{1}{c|}{$2^{nd}$ layer} & \multicolumn{1}{c|}{31.12} & {0.9880}  \\ \cline{3-5} 
                  &                   & \multicolumn{1}{c|}{$3^{rd}$ layer} & \multicolumn{1}{c|}{18.32} & {0.8353}  \\ \cline{3-5} 
                  &                   & \multicolumn{1}{c|}{$4^{th}$ layer} & \multicolumn{1}{c|}{14.36} & {0.6572}  \\ \cline{3-5} 
                  &                   & \multicolumn{1}{c|}{$5^{th}$ layer} & \multicolumn{1}{c|}{11.91} & {0.5719}  \\ \cline{3-5} 
                  &                   & \multicolumn{1}{c|}{$6^{th}$ layer} & \multicolumn{1}{c|}{10.79} & {0.5228}  \\ \cline{3-5} 
                  &                   & \multicolumn{1}{c|}{$7^{th}$ layer} & \multicolumn{1}{c|}{8.73} & {0.4246}  \\ \cline{3-5} 
                  &                   & \multicolumn{1}{c|}{$8^{th}$ layer} & \multicolumn{1}{c|}{7.92} & {0.3805}  \\ \cline{3-5} 
                  &                   & \multicolumn{1}{c|}{$9^{th}$ layer} & \multicolumn{1}{c|}{6.86} & {0.3208}  \\ \cline{3-5} 
                  &                   & \multicolumn{1}{c|}{$10^{th}$ layer} & \multicolumn{1}{c|}{4.83} & {0.1228}  \\ \hline
\end{tabular}
\label{Quantitative comparison using the PSNR and SSIM}
\end{table}

\section{Conclusion}
\label{section5}
In this paper, a new Deep Bi-directional Predictive Coding (DBPC) that supports developing networks that can simultaneously perform classification and reconstruction has been proposed and proven to operate efficiently. The proposed DBPC builds on existing PC methods by developing a network in which each layer simultaneously predicts the activities of neurons in the previous and next layer to simultaneously perform classification and reconstruction tasks. The performance of networks trained using DBPC has been evaluated for classification and reconstruction tasks using the MNIST and FashionMNIST datasets. The results of performance comparison clearly indicate that the classification and reconstruction performance of DBPC is similar to other existing approaches but, DBPC employs a significantly smaller network. In addition, DBPC relies on locally available information for learning and employs in-parallel updates across all layers in the network which results in a more efficient training protocol. Future directions on this work will focus on extending the reconstruction capabilities of DBPC to generate samples in the input space.

\bibliographystyle{IEEEtran}
\bibliography{bibliography.bib}
% \end{thebibliography}

\newpage

 

% \begin{IEEEbiography}[{\includegraphics[width=1in,height=1.25in,clip,keepaspectratio]{Template/fig1.png}}]{Michael Shell}
% \end{IEEEbiography}



\vfill

\end{document}